\documentclass{llncs}

\usepackage{graphicx}   
\usepackage{subfigure}

\begin{document}


\def\dlm{{\em DLM}}
\def\psc{{\em PRV}}

\def\pang{{\varphi_{y}}}
\def\tang{{\varphi_{x}}}
\def\vang{{\varphi_{z}}}
\def\sgn{{\rm{sgn}}}
\def\fref#1{(Fig.~\ref{#1})}
\def\sref#1{(Section~\ref{sec:#1})}

\def\backg{{\em background}\/ }
\def\foreg{{\em foreground}\/ }

\renewcommand{\labelitemi}{-}
\renewcommand{\labelitemii}{+}
\newcommand{\figend}{.pdf}
\newcommand{\figfile}[1]{#1.pdf}
\newenvironment{unterschr}[1]{
  \begin{center}\mbox{(a)}\hspace{#1}\mbox{(b)}\\}
  {\end{center}}
\newenvironment{my_figure}[2]{
  \begin{figure}[ht]\centerline{\setlength{\fboxsep}{0.5mm}
     \includegraphics[width=#2]{\figfile{#1}}
     }}
  {\end{figure}}
\newenvironment{dummy_figure}[2]{
  \begin{figure}[ht]\centerline{\setlength{\fboxsep}{2mm}\fbox{
     \rule{0cm}{#2}\rule{#1}{0cm}}}
     }
  {\end{figure}}
\newenvironment{text_fig}[4]{
  \begin{wrapfigure}{#2}{#3}
  \fbox{\psfig{file=\figfile{#1},width=#4}
   }}
  {\end{wrapfigure}}

\newenvironment{double_fig}[4]{
 \begin{figure}[ht]
 \begin{center}
  \setlength{\fboxsep}{0.5mm}
  \frame{\includegraphics[height=#2]{\figfile{#1}}}
  \hspace{0.1cm}
  \frame{\includegraphics[height=#4]{\figfile{#2}}}\\
  \begin{minipage}[h]{3.5cm}\begin{center}{\footnotesize(a)}\end{center}
        \end{minipage}\hspace{0.01cm}
  \begin{minipage}[h]{3.5cm}\begin{center}{\footnotesize(b)}\end{center}		\end{minipage}
 \end{center}}
{\end{figure}}

\newenvironment{doubl_fig}[3]{
 \begin{figure}[ht]
 \begin{center}
  \setlength{\fboxsep}{0.5mm}
  \includegraphics[height=#3]{\figfile{#1}}
  \hspace{0.01cm}
  \includegraphics[height=#3]{\figfile{#2}}
 \end{center}}
{\end{figure}}

\newenvironment{tripl_fig}[4]{
 \begin{figure}[ht]
 \begin{center}
  \setlength{\fboxsep}{0.5mm}
  \frame{\includegraphics[height=#4]{\figfile{#1}}}
  \hspace{0.1cm}
  \frame{\includegraphics[height=#4]{\figfile{#2}}}
  \hspace{0.1cm}
  \frame{\includegraphics[height=#4]{\figfile{#3}}}
 \end{center}}
{\end{figure}}

\newenvironment{tripl_title}[4]{
 \begin{figure}[ht]
 \begin{center}
  \psfig{file=\figfile{#1},height=#4}
  \hfill
  \psfig{file=\figfile{#2},height=#4}
  \hfill
  \psfig{file=\figfile{#3},height=#4}
 \end{center}}
{\end{figure}}

\newenvironment{quad_fig}[5]{
 \begin{figure}[ht]
 \begin{center}
  \setlength{\fboxsep}{0.5mm}
  \frame{\fbox{\psfig{file=\figfile{#1},height=#5}}}
  \hspace{0.1cm}
  \frame{\fbox{\psfig{file=\figfile{#2},height=#5}}}
  \hspace{0.1cm}
  \frame{\fbox{\psfig{file=\figfile{#3},height=#5}}}
  \hspace{0.1cm}
  \frame{\fbox{\psfig{file=\figfile{#4},height=#5}}}
 \end{center}}
{\end{figure}}

\newcommand{\vecpl}[2]
   {\left(\begin{array}{c} #1\\[1ex]#2\end{array}\right)}

\newcommand{\vecsp}[3]
   {\left(\begin{array}{c}#1\\[1.2ex]#2\\[1.2ex]#3\end{array}\right)}

\newcommand{\under}[2]
   {\begin{array}{c}\\#1\\[-0.2ex]#2\end{array}}

\def\eqdef{\stackrel{\text{def}}{=}}

\title{Direct Pose Estimation with a  Monocular Camera}

\author{ Darius Burschka and Elmar Mair}
\institute{Department of Informatics\\
	Technische Universit\"at M\"unchen, Germany\\
	\texttt{\{burschka$|$elmar.mair\}@mytum.de}}

%
%
%

\maketitle

\begin{abstract}

We present a direct method to calculate a 6DoF pose change of a
monocular camera for mobile navigation. The calculated pose is estimated
up to a constant unknown scale parameter that is kept constant over the
entire reconstruction process. This method allows a direct calculation
of the metric position and rotation without any necessity to fuse the
information in a probabilistic approach over longer frame sequence 
as it is the case in most currently used VSLAM approaches.  The
algorithm provides two novel aspects to the field of monocular
navigation. It allows a direct pose estimation without any a-priori
knowledge about the world directly from any two images and  it
provides a quality measure for the estimated motion parameters that
allows to fuse the resulting information in Kalman Filters.

We present the mathematical formulation of the approach together with
experimental validation on real scene images.
\end{abstract}

\section{Motivation}

Localization is an essential task in most applications of a mobile or a
manipulation system.  It can be subdivided into two categories of the
initial (global) localization and the relative localization. While the former
requires an identification of known reference structures in the camera
image to find the current pose relative to a known, a-priori
map~\cite{Thrun98a}, often it is merely necessary to
register correctly the relative position changes in consecutive image
frames.

Many localization approaches for indoor applications use
simplifications like assumptions about planarity of the imaged objects
in the scene or assume a restricted motion in the
ground plane of the floor that allows to derive the metric navigation
parameters from differences in the images using Image Jacobians in
vision-based control approaches. A true 6DoF localization requires a
significant computational effort to calculate the parameters while
solving an octal polynomial equation~\cite{Nister} or estimating the
pose with a Bayesian minimization approach utilizing intersections of
uncertainty ellipsoids to find the true position of the imaged points
from a longer sequence of images~\cite{Davison}. While the first
solution still requires a sampling to find the true solution of the
equation due to the high complexity of the problem, the second one can
calculate the result only after a  motion sequence with strongly varying
direction of motion of the camera that helps to reduce the uncertainty
about the position of the physical point. 

A common used solutions to this problem are structure from motion
approaches like the {\em eight point algorithm}~\cite{higgins} and its
derivatives using as few as five corresponding points between the two
images. Another approach is to use homographies to estimate the motion
parameters in case that the corresponding points all are on a planar
surface. These approaches provide a solution to the motion
parameters~(R,T - rotation matrix, translation vector) but are very
sensitive to the ill conditioned point configurations and to outliers.
Usually, the noise in the point detection does not allow to detect
ill-conditioned cases and the system needs to cope with wrong estimate
without any additional information about the covariance of the estimated
values.  Our method follows the idea of metric reconstruction from the
absolute conic~$\Omega_{\infty}$~\cite{zisserman} for the calculation of
the pose parameters for any point configuration providing correct
covariance values for all estimates. Our contribution is a robust
implementation of the algorithm suppressing outliers in the matches
between the two images using RANSAC and a correct calculation of
covariance values for the resulting pose estimation.

Since the projection in a monocular camera results in a loss of one
dimension, the estimation of the three-dimensional parameters in space
usually requires a metric reference to the surrounding world, a 3D
model, that is used to scale the result of the image processing back to
Cartesian coordinates.  Assuming that the projective geometry of the
camera is modeled by perspective projection~\cite{Horn86}, a point,
$^cP=(x, y, z)^T$, whose coordinates are expressed with respect to the
camera coordinate frame~c, will project onto the image plane with
coordinates $p=(u,\nu)^T$, given by

\begin{equation}
  \pi(^cP)=\vecpl{u}{v}=\frac{f}{z}\vecpl{x}{y}
  \label{proj::eq}
\end{equation}

\noindent
Points in the image correspond to an internal model of the environment
that is stored in 3D coordinates $^mP$.

Localization with a monocular camera system is in this case formulated
as estimation of the transformation matrix~$^cx_m$ such that an image
point~$p=(u,\nu)^T$ corresponds to an actual observation in the camera
image for any visible model point $^mP$.
\begin{equation} 
p=\pi(^cx_m(^mP))
\label{gentrans_eq}
\end{equation}

In many cases the initial position of the system is known or not
relevant, since the initial frame may define the starting position.  We
propose a navigation system that operates in the image space of the
camera for this domain of applications.  The system maintains the
correspondences between the ``model'' and the world based on a simple
landmark tracking since both the model and the perception are defined in
the same coordinate frame of the camera projection $p=(u,\nu)^T$.

The paper is structured as follows, in the following section we present
the algorithms used for the estimation of the 3D~rotation matrix from
the points on the "horizon" and the way how the tracked points are
classified if far away from the camera.  We describe the way the
system processes the image data to estimate the translational component
in all three degrees of freedom. This estimation is possible only up to
a scale.  In Section~\ref{results:sec} we evaluate the accuracy of the
system for different motion parameters.  We conclude with an evaluation
of the system performance and our future plans.

\section{Imaging Properties of a Monocular Camera}

\begin{figure}
\includegraphics[width=4cm]{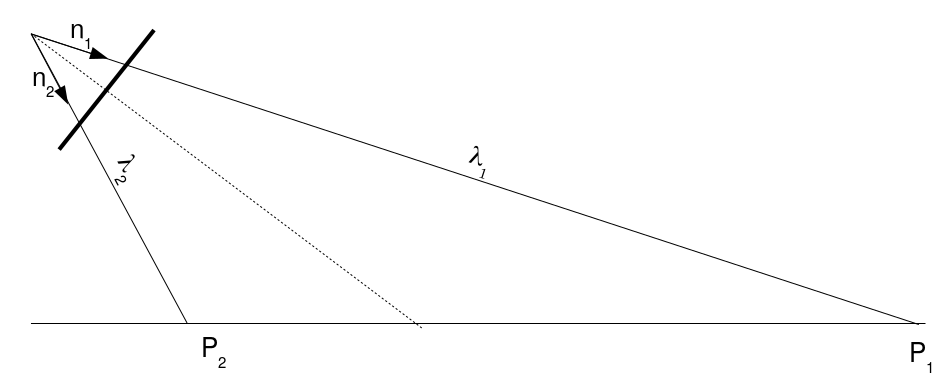}
\caption{\label{cam::fig} Radial distance~$\lambda_i$ is lost during the
camera projection. Only the direction~$\vec{n_i}$ to the imaged 
point can be obtained from the camera image.}
\end{figure}

As mentioned already above, a projection in a camera image results in
a reduction of the dimensionality of the scene. The information about
the radial distance to the point~$\lambda_i$ is lost in the projection
(Fig.~\ref{cam::fig}).

\noindent
Only the direction vector~$\overrightarrow {n_i}$ can be calculated from the 
camera projection to:
\begin{equation}
  \overrightarrow{k_i} = \vecsp{\frac{(u_{pi} - Cx)\cdot s_x}{f}}
  			      {\frac{(v_{pi} - Cy)\cdot s_y}{f}}
			      {1}\Rightarrow\quad
  \overrightarrow{n_i}=\frac{1}{||\overrightarrow{k_i}||}\overrightarrow{k_i}
  \label{direct:eq}
\end{equation}

\noindent with $(u_{pi},v_{pi})$ being the pixel coordinates of a point.
$(C_x,C_y)$ represent the position of the optical center in the camera
image, $(s_x,s_y)$ are the pixel sizes on the camera chip, and $f$ is
the focal length of the camera. This transformation removes all
dependencies from the calibration parameters, like the parameters of
the actual optical lens and its mount relative to the camera image. It
transforms the image information into a pure geometric line of sight
representation.

An arbitrary motion of the camera in space can be defined by the
3~translation parameters $\overrightarrow T=(X Y Z)^T$ and a rotation
matrix~R. The motion can be observed as an apparent motion of a point
from a position~$\overrightarrow P$ to a position~$\overrightarrow{P'}$

\begin{equation}
  \overrightarrow{P'}=R^T\overrightarrow P-\overrightarrow T
  \label{allg_trans:eq}
\end{equation}

Optical flow approaches have already proven that these two motions
result in specific shifts in the projected camera image. The resulting
transformation on the imaged points corresponds to a sum of the motions
induced by the rotation of the system~R and the consecutive
translation~$\overrightarrow T$.
Since we can write any point in the world according to
Fig.~\ref{cam::fig} as a product of its direction of view~$n_i$ and the
radial distance~$\lambda_i$, we can write (\ref{allg_trans:eq}) in the
following form
\begin{equation}
  \lambda_i'\overrightarrow{n_i'}=\lambda_iR^T\overrightarrow{n_i} -\overrightarrow T
  \label{spec_trans:eq}
\end{equation}

If we were able to neglect one of the influences on the point
projections then we should be able to estimate the single motion
parameters.

\subsection{Estimation of the Rotation}

There are several methods to estimate rotation of a camera from a set of
two images. Vanishing points can be used to estimate the rotation
matrix~\cite{zisserman}. The idea is to use points at infinity that do
not experience any image shifts due to translation. In our approach, we
use different way to estimate the rotation that converts the imaged
points points into their normalized direction vectors~$\vec{n_i}$
(Fig.~\ref{cam::fig}). This allows us a direct calculation of rotation
as described below and additionally provides an easy extension to
omnidirectional cameras, where the imaged points can lie on a
half-sphere around the focal point of the camera. We describe the
algorithm in more detail in the following text.

\begin{figure}
\includegraphics[width=6cm]{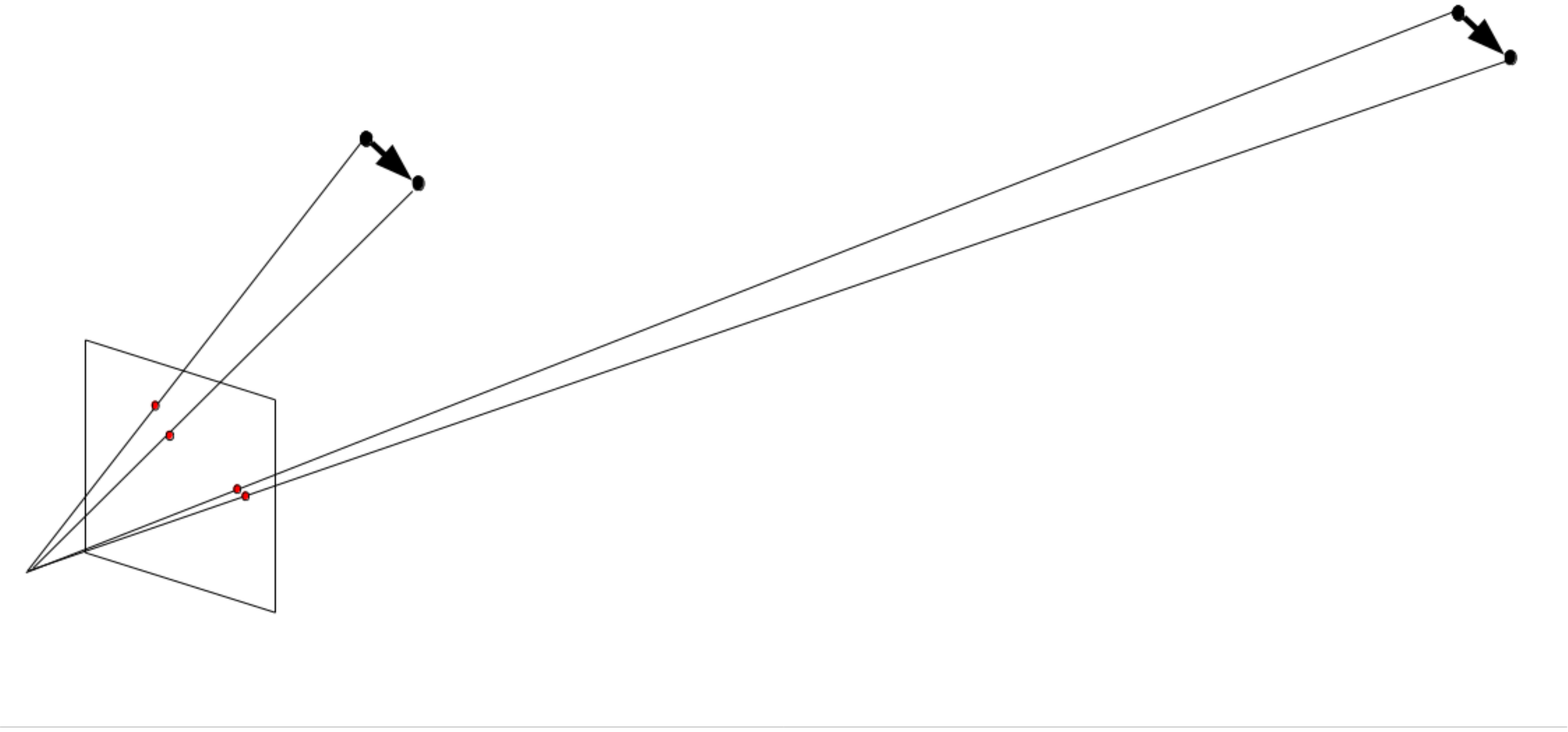}
\caption{\label{quantis:fig}The change in the imaged position for a
similar motion in 3D changes with the distance of the observed point to
the camera.}
\end{figure}

The camera quantizes the projected information with the resolution of
its chip~$(s_x,s_y)$. We can tell from the Eq.~(\ref{proj::eq}) that a
motion~$\Delta\overrightarrow T$ becomes less and less observable for large
values of~z. That means that distant points to the camera experience
only small shifts in their projected position due to the translational
motion of the camera (Fig.~\ref{quantis:fig}). 

\noindent The motion becomes not detectable once the resulting image
shift falls below the detection accuracy of the camera. Depending on the camera
chip resolution~$(s_x,s_y)$ and motion parameters between two
frames~$T_m$, we can define a set of such point~$P_{k\infty}$ in
the actual values as points in the distance~$Z_\infty$:

\begin{equation}
  Z_\infty=\frac{f}{s_x}T_m
  \label{negmotion:eq}
\end{equation}

We assumed in~(\ref{negmotion:eq}) that the pixels~$(s_x,s_y)$ are
square or the smaller of the both values needs to be taken into account.
Additionally, we assumed that the largest observed motion
component occurs for motions coplanar to the image plane and that the
feature detection works with the accuracy of up to one pixel. We see
that for a typical camera setup with a pixel size of~$s_x=11\mu m$ and a
focal length of~$f=8mm$ objects as close as 14m cannot be detected 
at motion speeds lower than~$T_m=2cm/frame$. This is a typical speed of a
mobile robot in the scene. Typically, this value is significantly smaller
because the motion component parallel to the image plane for a camera looking 
forward is smaller (Fig.~\ref{motion:fig})

\begin{my_figure}{motion}{3cm}
\caption{\label{motion:fig}Motion component~$T_m$ that can be observed in the
image.}
\end{my_figure}

\noindent
We can calculate the motion value~$T_m$ from the actual motion of the
vehicle~$\Delta T$ in Fig.~\ref{motion:fig} to
\begin{equation}
  T_m=\frac{\cos\gamma}{\cos\varphi}\Delta T
\end{equation}

\noindent
Therefore, the typical observed motion~$T_m$ is smaller than the value
assumed above for a motion observed by a camera looking parallel to the
motion vector~$\Delta T$. This would correspond to a camera looking to
the side. An important observation is that~$T_m$ is not only scaled by
the distance to the observed point but the angle~$\gamma$ is important
for the actual observability of a point motion. We see from this
equation directly that a radial motion as it is the case for points
along the optical center but also any other motion along the line of
projection lets a point render part of the $\{P_{k\infty}\}$ set of
points, from which the translation cannot be calculated.

We use this observation to subdivide the observed points in the camera
image into points which are in the distance larger than~$Z_\infty$
in~(\ref{negmotion:eq}) and those that are closer than this value.  We
use the Kanade-Lucas tracker (KLT) to track image points in a sequence
of images acquired by a monocular camera system.  The scene depicted in
Fig.~\ref{realscene:fig} contains both physical points~$\{P_{lT}\}$ that
let observe both motion types translation and rotation, and
points~$\{P_{k\infty}\}$ where we observe only the result of the
rotation.  Before we explain, how we decide which points belong to which
set, we make the observation that for the point set~$\{P_{k\infty}\}$
the equation~(\ref{spec_trans:eq}) simplifies to

\begin{equation}
\begin{array}{c}
 \forall\{P_{k\infty}\}:\quad\lambda_i'\overrightarrow{n_i'}=\lambda_iR^T\overrightarrow{n_i},
 \quad{\rm with}\quad \lambda_i'\simeq\lambda_i\\[2ex]
 \Rightarrow \quad\overrightarrow{n_i'}=R^T\overrightarrow{n_i}
 \end{array}
 \label{rotateq:ref}
\end{equation}

\begin{my_figure}{realrotation}{10cm}
\caption{\label{realscene:fig}We use Kanade-Lucas tracker (KLT) to track
points in a sequence of outdoor images.}
\end{my_figure}

It is known that the matrix~${\bf \tilde{R}}$ and the vector~${\bf T}$
between two 3D~point sets $\{P_i\}$ and $\{P^*_i\}$
can be recovered by solving the following least-square problem for {\em
N}~landmarks

\begin{equation}
\min_{{\bf\tilde{R}},{\bf T}}\sum_{i=1}^N \|{\bf\tilde{R}}P_i + {\bf T}
- P^*_i
\|^2,
\quad {\rm subject\: to\: } {\bf\tilde{R}}^T{\bf\tilde{R}} = I.
\label{eq:abs-ori-obj}
\end{equation}

\noindent
Such a constrained least squares problem can be solved in closed form using
quaternions \cite{AbsOri:quaternion-by-Horn,AbsOri:quaternion-by-WSV},
or singular value decomposition (SVD)
\cite{AbsOri:SVD-by-Horn,AbsOri:SVD-by-AHB,AbsOri:quaternion-by-Horn,AbsOri:quaternion-by-WSV}.
 
The SVD solution proceeds as follows.  Let $\{P_i\}$ and
$\{P^*_i\}$ denote lists of corresponding vectors  
and define
 \begin{equation}
 \bar{P} = \frac{1}{n} \sum_{i=1}^n P_i,
 \quad
 \bar{P^*} = \frac{1}{n}\sum_{i=1}^n P^*_i,
 \end{equation}
 that is, 
 $\bar{P}$ and $\bar{P^*}$ are the centroids of
 $\{P_i\}$ and $\{P^*_i\}$, respectively.
 Define
 \begin{equation}
 \label{mean_eq}
 P^\prime_i = P_i - \bar{P},
 \quad
 P'^*_i = P^*_i - \bar{P^*},
 \end{equation}
 and
 \begin{equation}
 \tilde{\bf M} = \sum_{i=1}^n P'^*_i{P^\prime_i}^{\bf T}\label{eq:M}.
 \end{equation}
 In other words, $\frac{1}{n} \tilde{\bf M}$ is the sample
cross-covariance matrix between $\{P_i\}$ and $\{P^*_i\}.$
It can be shown that, if $\tilde{\bf R}^*$, ${\bf T}^*$ minimize
(\ref{eq:abs-ori-obj}), then they satisfy
 \begin{equation}
 \tilde{\bf R}^* = {\rm arg max}_{\tilde{\bf R}} {\rm tr}(\tilde{\bf R}^T
\tilde{\bf M})\label{eq:rotmax}
 \label{transsol_eq}
 \end{equation}

\noindent
Since we subtract the mean
value~$\bar{P}$ in (\ref{mean_eq}), we remove any translation component
leaving just a pure rotation.

 Let $(\tilde{U},\tilde{\Sigma},\tilde{V})$ be a SVD of
$\tilde{\bf M}$. Then the solution to (\ref{eq:abs-ori-obj}) is
 \begin{equation}
 \tilde{\bf R}^* = \tilde{\bf V}\tilde{\bf U}^T
 \label{eq:SVDsol}
 \end{equation}

We use the solution in equation~(\ref{eq:SVDsol}) as a least-square
estimate for the rotation matrix~R in~(\ref{rotateq:ref}). The
3D~direction vectors $\{n_i\}$ and $\{n'_i\}$ are used as the
point sets~$\{P_i\}$ and $\{P^*_i\}$ in the approach described above. It
is important to use the~$\overrightarrow{n_i}$ vectors and not
the~$\overrightarrow{k_i}$ vectors from~(\ref{direct:eq}) here.

The reader may ask, how can we make this distinction which points belong
to $\{P_{k\infty}\}$ not knowing the real radial distances~$\lambda_i$
to the imaged points like the ones depicted in Fig.~\ref{realscene:fig}?
It is possible to use assumptions about the field of view of the camera
to answer this question, where points at the horizon appear in well
defined parts of the image.  We apply a generic solution to this problem
involving the Random Sample Consensus (RANSAC). It was introduced by
Bolles and Fischler~\cite{ransac}. RANSAC picks a minimal subset of 3
tracked points and calculates the least-square approximation of the
rotation matrix for them (\ref{eq:SVDsol}). It enlarges the set with
consistent data stepwise. We used following Lemma to pick  consistent
points for the calculation of~R.\\[2ex]

\noindent
{\it Lemma:}
All points used for the calculation of~R
belong to the set~$\{P_{k\infty}\}$, iff the calculated R warps the
appearance of the point features~$\{n_i\}$ to the appearance in the
second image~$\{n'_i\}$.\\

The resulting image error for correct points goes to zero after applying
the 3D~rotation matrix to their corresponding direction vectors. Note
that we estimate here directly the 3 rotational parameters of the camera
system and not the rotation in the image space like most other
approaches.

\begin{figure}[ht]
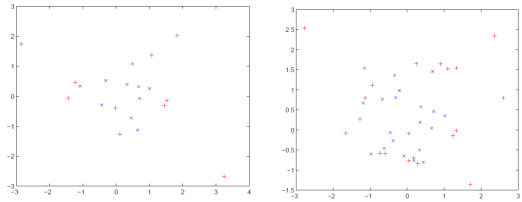

\begin{center}
\includegraphics[height=2.7cm]{\figfile{left}}
\hspace{0.1cm}
\includegraphics[height=2.7cm]{\figfile{right}}\\
\end{center}
\caption{\label{testsyn:fig} Two synthetic sets of feature positions with close 
(+) and
distant points (x) with a camera translation by $\Delta T=(0.3, 0.2, 0.4)^T$[m]
and a rotation around the axes by $(10^\circ,2^\circ,5^\circ)$.}
\end{figure}

We can validate it at the following example in Fig.~\ref{testsyn:fig}.
We generated a set of feature points in a close range to the camera in a
distance 1-4m in front of the camera. These points are marked with (+)
in the images. A second set of points was placed in a distance 25-35m in
front of the camera and these points are marked with (x).  We estimated
the rotational matrix from a set  of direction vectors~$n_i$ estimated
with RANSAC that matched the (x) points. These points represent
the~$\{P_{k\infty}\}$ set of our calculation. The calculated 3D~rotation
matrix had a remaining error of
$(-0.1215^\circ,0.2727^\circ,-0.0496^\circ)$.

\begin{figure}[ht]
\begin{center}
\includegraphics[height=3.5cm]{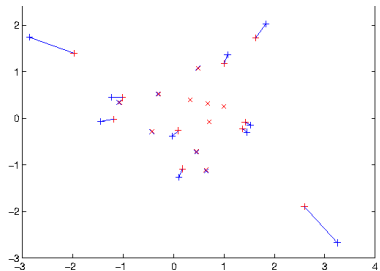}\\
\end{center}
\caption{\label{comrot:fig} Features from Fig.~\ref{testsyn:fig}
compensated for rotation estimated from (x) feature positions.}
\end{figure}

We used the rotation matrix to compensate all direction vectors in the
second image~$\{n_i'\}$ for the influence of rotation during the camera
motion. Fig.~\ref{comrot:fig} depicts the positions of all features
after compensation of the rotation influence. We see that as we already
expected above the distant features were moved to their original
positions. These features do not give us any information about the
translation. They should not be used in the following processing. The
close features (marked with '+') in Fig.~\ref{comrot:fig} still have a
residual displacement in the image after the compensation of the
rotation that gives us information about the translation between the two
images.

\subsection{Estimation of the Translation}
\label{trans:sec}

In the equation~(\ref{spec_trans:eq}), we identified two possible motion
components in the images. It is known that the 3D~rotation of the camera
relative to the world can always be calculated, but the distances to the
imaged points can only be calculated once a significant motion between
the frames is present (see Fig.~\ref{comrot:fig}). We need some
remaining points with significant displacement between the images (after
they were compensated for rotation) to estimate the translation
parameters. We estimate the translation of the camera from the position
of the epipoles.
\begin{my_figure}{translation}{5cm}
\caption{\label{transl:fig} In case that the camera is moved by a pure
translation, the direction of the epipole~$T$ defines the direction of
the translational motion.}
\end{my_figure}

We know from Fig.~\ref{cam::fig} that the camera measures only the
direction~$n_i$ to the imaged point but looses the radial
distance~$\lambda_i$.  Therefore, we know only the direction to both
points~$(P_1,P_2)$ for the left camera position in
Fig.~\ref{transl:fig}.  If we find correspondences in both cameras then
we know that the corresponding point lies somewhere on a line starting
at the projection of the focal point of the right camera in the left
image (epipole) and that it intersects the image plane of the right camera in the
corresponding point. The projection of this line in the left image is
called the epipolar line. We know that both corresponding points lie on
the epipolar line and that the epipole defines the beginning of the line
segment. Therefore, it is possible to find the position of the epipole
by intersecting two such lines since all epipolar lines share the
same epipole. Since the points~$(P_i,F_1,F_2)$ define a plane, all
possible projections of the given point while travelling along the line
$\overline{F_1F_2}$ will appear on the epipolar line.

We need at least two corresponding point pairs in the~$\{P_{lT}\}$  to
estimate the position of the epipole in the image. For each of the line
pairs, we calculate the following equation using the 2D
projections~$\overrightarrow{k_{ip}}$ of the
$\overrightarrow{k_i}$ vectors from the equation~(\ref{direct:eq}) on
the image plane. As simplification, we write in the following
equation~(\ref{intersect:eq})
$\overrightarrow{k_i}$ for them, but we mean the 2D~projections.
\begin{equation}
   \begin{array}{c}
     \overrightarrow{k_a}+\mu*(\overrightarrow{k'_a}-\overrightarrow{k_a})=
     \overrightarrow{k_b}+\nu*(\overrightarrow{k'_b}-\overrightarrow{k_b})\\
     \vecpl{\mu}{\nu}=\left(
        \overrightarrow{k_a}-\overrightarrow{k'_a} \quad
	\overrightarrow{k'_b}-\overrightarrow{k_b}\right)^{-1}\cdot
	\left(\overrightarrow{k_a}-\overrightarrow{k_b}\right)
   \end{array}
   \label{intersect:eq}
\end{equation}

\noindent
We can estimate the image position of the epipole~$E_p$ from the estimated
values~$(\mu, \nu)$ using the 3D versions of the
vectors~$\overrightarrow{k_i}$ to for example:
\begin{equation}
   \overrightarrow{E_p}=\overrightarrow{k_a}+\mu*(\overrightarrow{k'_a}-\overrightarrow{k_a})
\end{equation}

\noindent Note that the position of the epipole is identical for both images after
the compensation of rotation. The epipole defines the
direction of motion between the two cameras as depicted in
Fig.~\ref{transl:fig}. It is a known fact that we can reconstruct the
motion parameters only up to an unknown scale from a monocular camera.
Therefore, all we can rely on is the direction of motion
\begin{equation}
  \overrightarrow T=\frac{c}{|| \overrightarrow{E_p}||}
  \overrightarrow{E_p}
  \label{motion:eq}
\end{equation}

\noindent
The value~c in~(\ref{motion:eq}) can have a value of~$\pm1$ depending on
the direction in which the corresponding points move in the images. If
the corresponding point moves away from the epipole then we assume c=1
and if the corresponding point moves toward the epipole then we assume
c=-1. This explains, why we get two identical epipoles for both
translated images. These images have similar translation vectors. The
only difference is the sign (c-value).

For the example from Fig.~\ref{comrot:fig}, we calculate the position of
the epipole to be $\overrightarrow{[E_p}=(0.7, 0.36, 1)^T$ that matches
exactly the direction of the motion that we used to create the data. We
can get multiple solutions for the position of the epipole in the image.
The different
results originate from intersection of different line lengths, which
allows to pick the lines with the longest image extension to reduce the
detection error and increase the accuracy. A better option is to use a
weighted~$\omega_i$ average of the estimated values~${E_p}_i$. The
weight~$\omega_i$ depends on the minimal length~$l_i$ of the both line
segments and it is calculated to:
\begin{equation}
 \omega_i=\left\{ \begin{array}{ll}
         \frac{l_i}{L},&li<L\\
	 1,& li\ge L\end{array}\right.\Rightarrow\quad
 \overrightarrow{E_p}=\frac{\sum_i (\omega_i{\overrightarrow{E_p}}_i)}{\sum_i \omega_i}
\end{equation}

\indent
The value~L is chosen depending on the actual accuracy of the feature
detection. We set it to L=12~pixels in our system, which means that
displacements larger than 12~pixels for translation will be considered
accurate.

\subsection{Calculation of the Translational Error}

We need to estimate the covariance of the resulting pose estimation
directly from the quality of the data in the images. It is obvious that
accuracy increases with the increasing length of the image vectors of
the optical flow. In these cases, detection errors have smaller
influence. In general, the vectors of the translational field do not
intersect in one point (the epipole) as in the ideal case, but they
intersect in an area around the epipole. The area gets smaller if the
angle between the vectors is close to~$90^\circ$ and increases the more
the vectors become parallel to each other. Parallel vectors intersect in
infinity and does not allow any robust motion estimation in our system.

We can estimate the variance of the resulting translational vector from
the variance of the intersecting point vector pairs of the translational
vector field. It is very important to provide this value as an output of
the system, because calculated translational value may be
ill-conditioned due to very short almost parallel vectors that may
intersect in a large area around the actual epipole position.

\section{Results}
\label{results:sec}

We applied the presented system on a variety of images in indoor and
outdoor scenarios. The system was able to calculate both the rotation
matrix~R in all cases and the direction of the translation depending on
the matching criterions in the RANSAC process calculating the inliers
(points used for rotation) and outliers based on compensation of the
rotation and evaluation of the resulting re-projection error. This
matching can allow deviations of the position of the reprojected points.
In case of poor camera calibration and/or poor feature detection the
matching must be set to higher values allowing larger variations. This
requires automatically a larger translations between the frames.

We validated the system on synthetic and real images to estimate the
accuracy of the system.

\subsection{Simulation Results}

We tried to estimate the accuracy of the algorithm assuming a sub-pixel
accuracy in point detection in the images. We added white noise to the
ideal values with~$\sigma^2=0.05$.  We added 20 outliers to the matches
provided to the algorithm. We validated it on 2000 images. We observed
following accuracies:

\begin{table}
        \begin{center}
          \begin{tabular}{|l|r|}
                        \hline
                        average number of  matches&
89.7725 \\
                        average rotation error around the
x-axis & $0.0113^\circ$\\
                        average rotation error around the
y-axis & $0.0107^\circ$ \\
                        average rotation error around the
z-axis & $0.0231^\circ$ \\
                        number of failed estimates of rotation & 0
from 2000 \\
                        number of failed estimates of translation 
& 0 from 2000 \\
                        average error in the direction of the estimated
translation vector & $2.5888^\circ$ \\
                        \hline
          \end{tabular}
        \end{center}
\end{table}

The system provides robust estimates of the pose even under presence of
outliers, which is an important extension to the typical structure from
motion algorithm.

\subsection{Tests on Real Images}

We tested the system using a calibrated digital camera that we used to
acquire a sequence of images in different scenarios. We ran a
C++-implementation of the KLT-tracker on this sequence of images that
created a list of tracked positions for single image features. We used
this lists in Matlab to evaluate the rotation and translation
parameters.
\begin{figure}
\includegraphics[width=10cm]{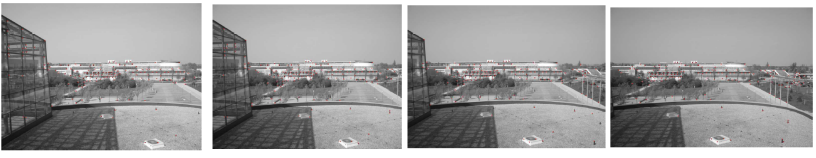}
\caption{\label{seq1:fig} Horizontal sweep of the camera with mostly
rotation.}
\end{figure}

\noindent
For the sequence in Fig.~\ref{seq1:fig}, we obtained the following
results:\\
\begin{center}
\begin{tabular}[h]{|p{4cm}|p{2cm}|p{2cm}|p{2cm}|}
\hline
rotation $(\alpha,\beta,\gamma)$ in [$^\circ$] & inlier&outlier&error\\
\hline
\hline
(2.132,-0.077,0.108) & 49 & 5 & 0.003066\\
(4.521,-0.007,0.0751)& 11 & 7 & 0.003874\\
(1.9466,0.288,-0.071) & 56 & 6 & 0.004\\
\hline
\end{tabular}
\end{center}
\begin{figure}
\includegraphics[width=10cm]{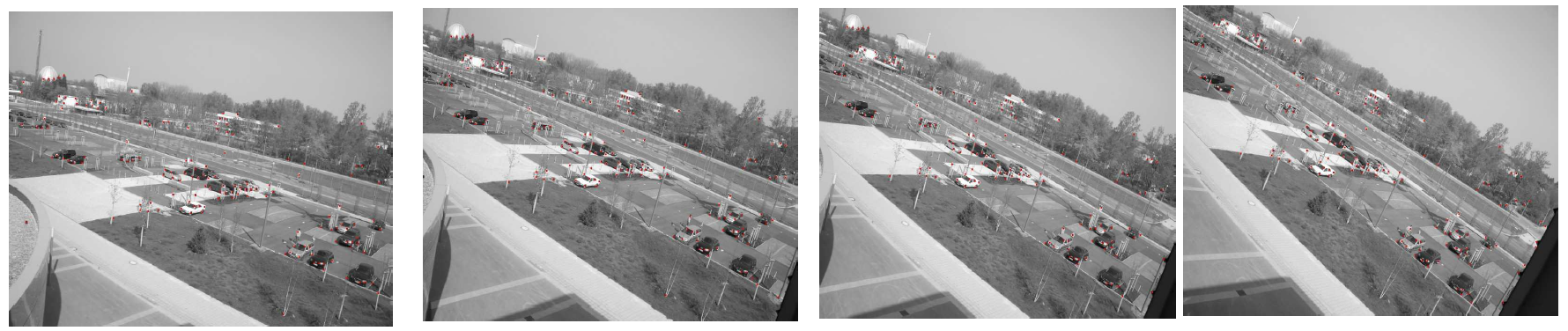}
\caption{\label{seq2:fig} Rotation around several axes simultaneously.}
\end{figure}
 We tested the estimation of significant rotation around all axes in
 Fig.~\ref{seq2:fig} with the following results:\\
\begin{center}
\begin{tabular}[h]{|p{4cm}|p{2cm}|p{2cm}|p{2cm}|}
\hline
rotation $(\alpha,\beta,\gamma)$ in [$^\circ$] & inlier&outlier&error\\
\hline
\hline
(1.736,-2.3955,6.05) & 24 & 1 & 0.0024\\
(1.08,0.64,4.257)& 87 & 0 & 0.0012\\
(2.2201,-0.2934,6.7948) & 56 & 6 & 0.004\\
\hline
\end{tabular}
\end{center}

The second sequence in Fig.~\ref{seq2:fig} shows the ability of the
system to capture the rotation of the camera correctly in all three
possible rotational degrees of freedom in each frame without any iterations
necessary. 
Due to the page limit, we cannot show our further results in indoor
scenarios, where we were able to provide some significant translation
that allows a simultaneous estimation of the translation direction.

We tested the system on real images taken from a car driven on our
campus and through the city of Garching close to Munich.
\begin{figure}[ht]
\begin{center}
\includegraphics[width=0.8\textwidth]{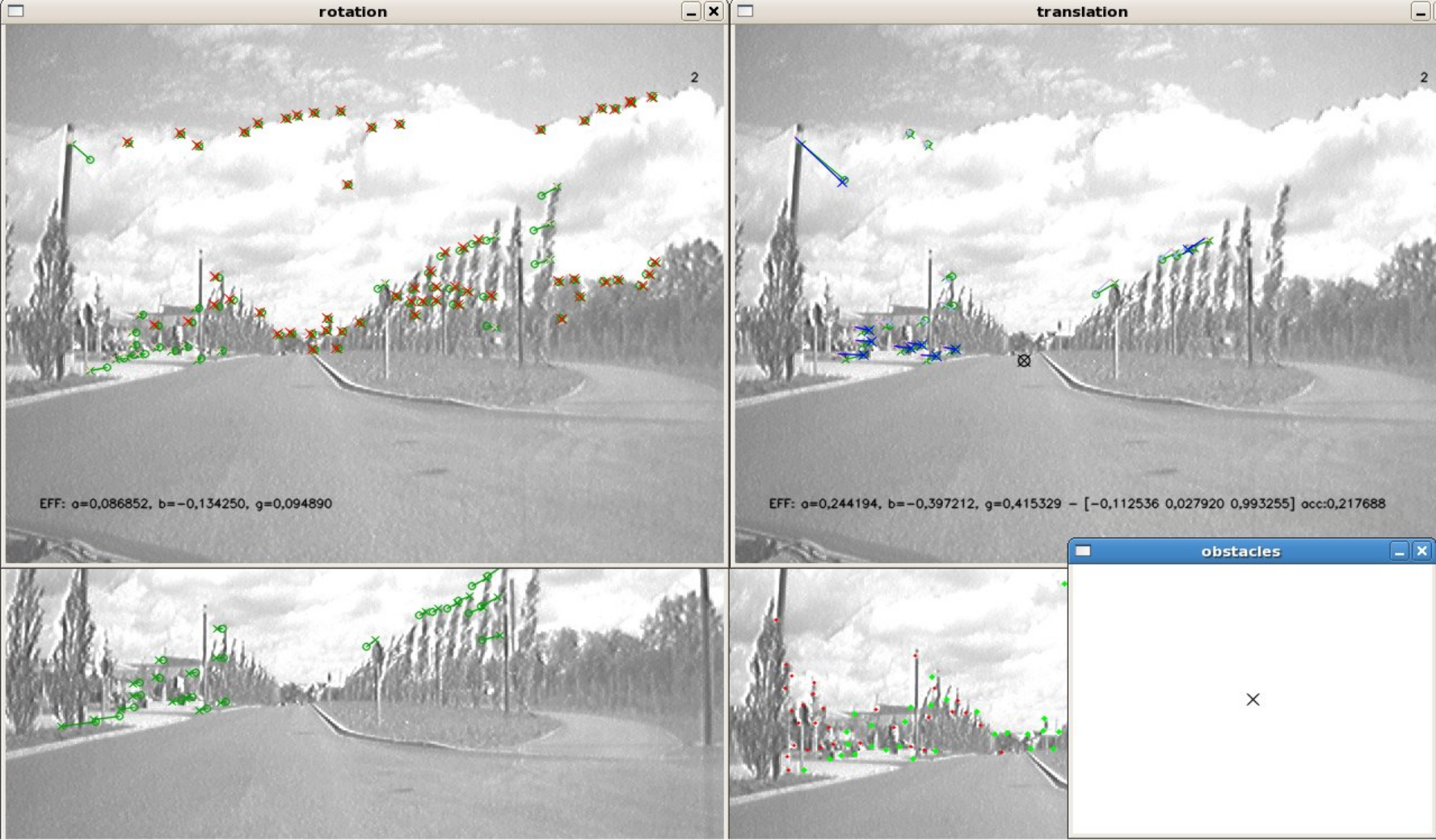}\\
\vspace{0.2cm}
\includegraphics[width=0.8\textwidth]{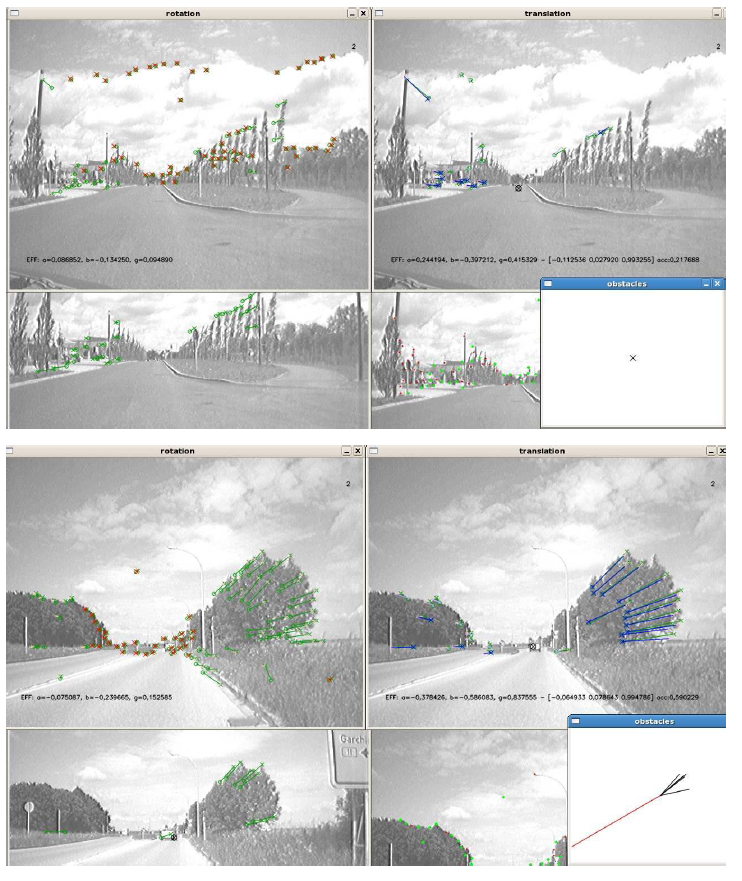}\\
        \end{center}
        \caption[Tests on real image]{Shown are two screen-shots of our
working navigation system showing the results for two scene examples
(top) drive through the Garching campus, (bottom) drive through the city
of Garching.\label{ref:visualisierung}}
\end{figure}

Fig.~\ref{ref:visualisierung} depicts the results of real navigation in
a city and campus scenario. The red points in the top left images are
the points used to estimate the rotation parameters while the green
points are the points used to estimate the translation. The top right
image shows the resulting translational optical flow and the position of
the estimated epipole as a black circle with a cross. The estimated
motion parameters with accuracies are also shown in
Fig.~\ref{ref:visualisierung}.

\vspace{-2ex}
\section{Conclusions and Future Work}

We presented a way to calculate pose and covariance values for
camera-based navigation with a monocular camera. The approach calculates
directly the 6DoF motion parameters from a pair of images and estimates
the accuracy for the estimated values that allows a correct fusion of
the resulting values in higher level fusion algorithms used for SLAM,
like Kalman Filters.  We provide the mathematical framework that allows
to subdivide tracked feature points into two sets of points that are close
to the sensor and whose position change in the image due to a combined
rotation and translation motion, and points that are far away and
experience only changes due to the rotation of the camera.  The
presented system was tested on synthetic and real world images proving
the validity of this concept.

The system is able to calculate the 6DoF parameters in every camera step
without any necessity for initialization of scene structure or initial
filtering of uncertainties.  In the next step, we will test the
translation estimates using external references for motion of the real
camera system and integrate the sensor into our VSLAM system.

\vspace{-2ex}
\bibliographystyle{plain}
\bibliography{./darius}
\end{document}